\documentclass[review]{elsarticle}

\usepackage{lineno,hyperref}
\usepackage{soul}
\usepackage{url}
\usepackage{graphicx}
\usepackage{amsmath}
\usepackage{amsthm}
\usepackage{booktabs}
\usepackage{algorithm}
\usepackage{algorithmic}
\usepackage{todonotes}
\usepackage{epsfig}
\usepackage{amssymb}
\usepackage{tabularx}%,booktabs,blindtext}
\usepackage{multirow, makecell}
\usepackage{array}
\usepackage{adjustbox}
\usepackage{rotating}

\modulolinenumbers[5]

\journal{Journal of \LaTeX\ Templates}

\begin{document}

\begin{frontmatter}

\title{Transductive Zero-Shot Learning using Cross-Modal CycleGAN}
%\tnotetext[mytitlenote]{Fully documented templates are available in the elsarticle package on \href{http://www.ctan.org/tex-archive/macros/latex/contrib/elsarticle}{CTAN}.}

%% Group authors per affiliation:
\author{Patrick Bordes$^1$, Eloi Zablocki$^2$, Benjamin Piwowarski$^1$, Patrick Gallinari$^{1,3}$}
\address{1: Laboratoire d'Informatique de Paris 6, 4 Place Jussieu, Paris, France}
\address{2: Valeo.ai, 15 rue de la Baume, Paris, France}
\address{3: Criteo AI Lab, 32 rue Blanche, Paris, France}
%\fntext[myfootnote]{Since 1880.}

%% or include affiliations in footnotes:
%\author[mymainaddress,mysecondaryaddress]{Elsevier Inc}
%\ead[url]{www.elsevier.com}

%\author[mysecondaryaddress]{Global Customer Service\corref{mycorrespondingauthor}}
\cortext[mycorrespondingauthor]{Corresponding author: Patrick Bordes }
\ead{patrick.bordes@lip6.fr}

%\address[mymainaddress]{1600 John F Kennedy Boulevard, Philadelphia}
%\address[mysecondaryaddress]{360 Park Avenue South, New York}

\begin{abstract}
In Computer Vision, Zero-Shot Learning (ZSL) aims at classifying \emph{unseen} classes --- classes for which no matching training image exists.  
Most of ZSL works learn a cross-modal mapping between images and class labels for seen classes. 
However, the data distribution of seen and unseen classes might differ, causing a \textit{domain shift} problem. 
Following this observation, transductive ZSL (T-ZSL) assumes that unseen classes and their associated images are known during training, but not their correspondence. As current T-ZSL approaches do not scale efficiently when the number of seen classes is high, we tackle this problem with a new model for T-ZSL based upon CycleGAN. 
Our model jointly (i) projects images on their \textit{seen} class labels with a supervised objective and (ii) aligns \textit{unseen} class labels and visual exemplars with adversarial and cycle-consistency objectives. 
We show the efficiency of our Cross-Modal CycleGAN model (CM-GAN) on the ImageNet T-ZSL task where we obtain state-of-the-art results. We further validate CM-GAN on a language grounding task, and on a new task that we propose: \textit{zero-shot sentence-to-image matching} on MS COCO. 
\end{abstract}


\end{frontmatter}

%\linenumbers

%%%%%%%%% BODY TEXT
\section{Introduction}

Over the last decade, the exponential evolution of computing power, combined with the creation of large-scale image datasets such as ImageNet \cite{imagenet}, have enabled Convolutional Neural Networks %\cite{Lecun98gradient-basedlearning} 
to keep improving their performances% in the last decade
, with recent improvements  \cite{DBLP:conf/aaai/RealAHL19}
%\cite{DBLP:journals/corr/ZophVSL17,DBLP:conf/aaai/RealAHL19}
pushing forward the classification score every year. However, as noted in \cite{devise}, such models have several drawbacks: they are unable to make predictions that fall outside of the set of training classes, their training requires a large number of examples for each class, and despite outmatching humans on the ImageNet challenge, they are unable to mimic the human capacity to \textit{generalize} prior knowledge to recognize new classes.

To cope with these limitations, \textit{Zero-Shot Learning} \cite{DBLP:conf/nips/PalatucciPHM09} approaches  %\cite{devise,DBLP:conf/icml/ZablockiBSPG19} %\cite{DBLP:conf/cvpr/FarhadiEHF09,DBLP:conf/eccv/MensinkVPC12,devise,DBLP:journals/corr/FuYHXG15,DBLP:conf/icml/ZablockiBSPG19} 
have been proposed. In ZSL, two sets of classes are distinguished: the \textit{seen} classes, for which examples are available during training, and the \textit{unseen} classes, for which training images are not available. 
The information learned using seen classes can be generalized to unseen ones by leveraging auxiliary knowledge, which semantically relates seen and unseen classes, e.g. attributes \cite{DBLP:conf/iccv/ParikhG11}% \cite{DBLP:conf/nips/FerrariZ07,DBLP:conf/iccv/ParikhG11,zsldatasetawa} 
, or textual embeddings of class labels \cite{devise}. 
Evaluation is then carried out on the \textit{unseen} classes. 
%Following the development of Generative Adversarial Networks \cite{DBLP:conf/nips/GoodfellowPMXWOCB14} in AI, ZSL has used generative models either for data augmentation \cite{DBLP:conf/iccvw/JurieBH17} to create visual exemplars for unseen classes, or (2) to model class labels as distributions in a latent space \cite{DBLP:conf/aaai/WangPVFZCRC18} that are used as priors in conditional Variational Auto-Encoders \cite{DBLP:journals/corr/KingmaW13}. 
%\cite{zsl,devise}. 
%The key of ZSL is to use auxiliary knowledge to semantically relate classes from the seen and unseen classes; thus, class labels are embedded in a a common semantic representation space. 
However, all these approaches suffer from a \textit{domain shift} problem \cite{DBLP:journals/pami/FuHXG15}: since the sets of seen and unseen classes are disjoint and potentially unrelated, the learned projection functions are biased towards seen classes. 

To alleviate the domain shift problem, 
%(1) data augmentation can be performed using GANs \cite{DBLP:conf/iccvw/JurieBH17} to create visual exemplars for unseen classes %\cite{DBLP:conf/iccvw/JurieBH17, DBLP:conf/cvpr/XianLSA18,DBLP:journals/corr/abs-1905-04511} to generate visual exemplars for unseen classes so that a supervised classification model can be trained, 
%(2) generative approaches \cite{DBLP:conf/aaai/WangPVFZCRC18} % \cite{DBLP:conf/aaai/WangPVFZCRC18,DBLP:conf/cvpr/MishraRMM18,DBLP:conf/cvpr/VermaAMR18,DBLP:journals/corr/abs-1906-03038} 
%where classes are represented by distributions that are used as priors in a Variational Auto-Encoder \cite{DBLP:journals/corr/KingmaW13}, %so that the class maximizing the VAE lower bound is the predicted class at inference, 
%and (3) 
transductive ZSL (T-ZSL) has been proposed  \cite{DBLP:conf/cvpr/YeG17}. %DBLP:journals/corr/abs-1901-01570
In this setting, unseen class labels and their corresponding images are known at training time, but not their correspondence. 
Whereas current T-ZSL approaches take advantage of the geometric structure of the visual space, e.g, with K-means clustering \cite{DBLP:journals/corr/abs-1901-01570} or nearest neighbor search \cite{DBLP:conf/nips/ZhaoDGLXW18}, such methods do not scale to a high number of unseen classes, as we show experimentally in \autoref{expes}. Indeed, when the number of unseen classes is high (e.g. 20K unseen classes for ImageNet), the visual space is not well-clustered, and class distributions have substantial overlap. %; moreover, when using Word2vec embeddings as class labels representations, hubness problems arise \cite{DBLP:journals/corr/DinuB14}.
%,  due to complex overlapping class distributions and hubness problem . 

In the present paper, we tackle this problem by building upon CycleGan \cite{cyclegan}: image and label distributions are aligned with an adversarial loss while ensuring that the structure of both spaces is preserved with a cycle-consistency loss. Thus, useful information is learned from the unseen classes distribution without relying on the clustering of the visual space. Our work follows unsupervised translation works that proved efficient on large-scale uni-modal datasets \cite{DBLP:conf/iclr/LampleCRDJ18}, 
and extends them to a cross-modal setting where (P1) the geometry of the visual and textual spaces are essentially different and (P2) there is a high imbalance between the number of images and the number of class labels. 
In the model that we propose, called Cross-Modal CycleGAN (CM-GAN), two cross-modal mappings (text-to-image and image-to-text) are learned between a visual space and a textual space using: (i) a \textit{supervised} max-margin triplet objective trained on seen classes, (ii) an \textit{unsupervised} CycleGAN objective trained on unseen classes. %, composed of adversarial losses that align image and label distributions without any supervision, and of a cycle-consistency loss that ensures that mappings are invertible. 
We tackle (P1) by representing images as linear combinations of class labels representations, following the CONSE \cite{DBLP:journals/corr/NorouziMBSSFCD13} approach --- which enables textual and visual distributions to be close and thus adversarial losses to be learned --- and (P2) by using an alternating optimization scheme where we progressively refine the projection of images and labels.
% We tackle the unbalance between the number of visual and textual instances (P2) by proposing an alternating optimization scheme.
% so that images are intuitively "homogeneous" to word embeddings.
%As adversarial losses are often used between two spaces of similar nature  --- either image-to-image \cite{cyclegan} or text-to-text \cite{DBLP:conf/iclr/LampleCRDJ18} unsupervised translation --- we propose to use CONSE \cite{DBLP:journals/corr/NorouziMBSSFCD13} to represent images so that images are intuitively "homogeneous" to word embeddings. Indeed, in CONSE, images are represented as convex combinations of the top class label's embeddings  predicted by a CNN. Using such visual representations enables (1) visual and textual spaces to be semantically c \textit{zero-shot sentence-to-image matching}lose --- we show this result using the $mNNO$ metric defined in \cite{DBLP:conf/acl/CollellM18} and (2) to efficiently train CycleGAN objective between these two spaces and thus benefit from the unpaired information of the unseen classes to learn cross-modal mappings --- we show experimentally that CycleGAN objective captures useful information, which combined to a supervised loss trained on seen classes lead to state-of-the-art results. 

Our approach is entirely new to ZSL. First, it differs from standard generative approaches (inductive or transductive) \cite{DBLP:journals/corr/abs-1906-03038,DBLP:conf/cvpr/HuangWYW19}, which learn to generate visual exemplars for unseen data, in order to feed them to a supervised classifier. Indeed, in our model, CycleGAN is used to learn cross-modal mappings directly, taking inspiration from uni-modal unsupervised translation models \citep{DBLP:conf/iclr/LampleCRDJ18,cyclegan}. Second, it is different from standard transductive models, that only tackle attribute datasets with a low number of unseen classes, and that rely on the clusterization of the visual embedding space, as in \cite{DBLP:journals/corr/abs-1901-01570} where a K-means algorithm is learned on visual features of unseen data.

We show that CM-GAN is particularly successful when the number of unseen classes is high (\autoref{expes}), namely on the ImageNet T-ZSL task where it achieves state-of-the-art results.
%Our contributions are the following:
%(1) we propose a new approach (CM-GAN) to T-ZSL based on CycleGAN, %the effectiveness of CM-GAN on the ImageNet ZSL task, on which we provide state-of-the-art results,
We further validate the efficiency of CM-GAN on a language grounding task (\autoref{grounding}) and on a new task, namely \textit{zero-shot image-to-sentence matching}, on MS COCO (\autoref{section_mscoco}). %and on a language grounding task (Section \ref{grounding}). %demonstrate its effectiveness on MS COCO an est d ImageNet (section XXX), where we provide SOTA results, and further validate CM-GAN on a language grounding task (section XXX).

%(3) we further validate CM-GAN on a language grounding task. % that CM-GAN can enhance textual representations using both supervised and unsupervised visual information. 

\section{Related Work}

\paragraph{Zero-Shot Learning} 
The usual procedure in ZSL \cite{devise} %(before the arise of GANs in ZSL, see Subsection \ref{gans_in_zsl}) 
consists in (1) learning a mapping between the visual and the textual space %(or to a common space)
%a joint representation space 
so that images and class labels can be semantically related, (2) performing a nearest neighbor search to find the closest unseen class corresponding to a projected image. 
Pioneering works focused on hand-crafted attributes for the textual space \cite{DBLP:conf/iccv/ParikhG11} e.g, `IsBlack', `HasClaws'. 
%\cite{DBLP:conf/nips/FerrariZ07,DBLP:conf/iccv/ParikhG11,zsldatasetawa}
Since this involves costly and error-prone human labeling, most current works use word vector spaces \cite{DBLP:journals/corr/NorouziMBSSFCD13}, %\cite{zsl,devise} 
such as Word2Vec \cite{DBLP:conf/nips/MikolovSCCD13}, which do not suffer from these limitations.% or Glove \cite{glove}
%; concerning images,
%most ZSL works use CNNs to embed them in a visual embedding space %\cite{fu14,DBLP:conf/cvpr/AkataRWLS15,DBLP:conf/cvpr/FuS16,DBLP:conf/eccv/BucherHJ16,DBLP:conf/icml/Romera-ParedesT15,DBLP:conf/iccv/ZhangS15a,DBLP:journals/pami/LampertNH14} 
%by taking the penultimate layer. 

%The vast majority of ZSL works \cite{DBLP:conf/cvpr/AkataPHS13, DBLP:journals/pami/LampertNH14,DBLP:conf/iccv/ZhangS15a} are evaluated on attribute datasets, namely AWA1 \cite{zsldatasetawa}, AWA2 \cite{DBLP:journals/pami/XianLSA19}, CUB \cite{WelinderEtal2010}, SUN \cite{Patterson2012SunAttributes}, aPY \cite{zsldatasetawa}. In these datasets, images are manually annotated given a set of pre-defined attributes, and class vectors are thus derived from these manual annotations; and the total number of classes (both seen and unseen) is relatively small (espectively 50,50,200,717 and 32). In this paper, we tackle the more challenging ImageNet dataset, which contains 14M images and about 20K unseen classes. 

The first ZSL approaches learned a linear projection of visual features in the textual space, like DeViSE \cite{devise}.  % - we use the idea of the max-margin loss in our model to exploit the information of source classes; 
Other models, like CONSE \cite{DBLP:journals/corr/NorouziMBSSFCD13} or SYNC \cite{DBLP:conf/cvpr/ChangpinyoCGS16}, express an image as a mixture of other classes features (\textit{hybrid} models) --- the CONSE model is presented in \autoref{conse}. 
%Then, CONSE \cite{DBLP:journals/corr/NorouziMBSSFCD13} proposed to embed images as a convex combination of the word embeddings of the top classes retrieved by a CNN. Going further than CONSE, SYNC \cite{DBLP:conf/cvpr/ChangpinyoCGS16} adopts the point of view of \textit{manifold learning}: classifiers for unseen classes are built by combining classifiers of \textit{phantom} classes, that are embedded both in the semantic space and the model space. %in the model space, seen and phantom classes form a weighted bipartite graph;
%(4) SS-Voc \cite{DBLP:conf/cvpr/FuS16}: semi-supervised vocabulary-informed learning: decision boundaries are learned to separate classes by using distance constraints between seen and unseen vocabulary atoms;
%(5) DEM \cite{DBLP:conf/cvpr/ZhangXG17}: a deep ZSL model is trained to project semantic representations of classes in the visual space, which is used as a joint embedding space to alleviate the hubness problem usually encountered in the semantic space.
More recently, non-linear relations between modalities are investigated, as in EXEM \cite{DBLP:conf/iccv/ChangpinyoCS17} where a kernel-based regressor is learned. %maps semantic representations to visual exemplars while ensuring that the semantic space is clustered efficiently. 
Another alternative is the exploitation of the WordNet knowledge graph of ImageNet synsets as  \cite{DBLP:journals/corr/abs-1805-11724}; % --- %\cite{DBLP:conf/cvpr/0004YG18,DBLP:journals/corr/abs-1805-11724} 
%using Graph Convolutional Networks \cite{DBLP:journals/corr/BrunaZSL13} ---  % ,DBLP:conf/nips/DefferrardBV16,DBLP:conf/iclr/KipfW17}. 
%we do not compare with such models as we do not use the WordNet graph. % information. 
such methods are orthogonal to our work as they are based on complementary knowledge.

%\textbf{VZSL} \cite{DBLP:conf/aaai/WangPVFZCRC18} --- classes are represented by class-specific latent space distribution, that are used as prior in a supervised VAE trained on seen classes; at inference, the class maximizing the VAE lower bound is the predicted class;
%\textbf{CVAE-ZSL} \cite{DBLP:conf/cvpr/MishraRMM18}: the ZSL problem is seen as a missing data problem. The probability distribution of the visual features conditioned to the class representation is learned using a conditional VAE;
%\textbf{SE-ZSL}  \cite{DBLP:conf/cvpr/VermaAMR18}: using a VAE to generate novel exemplars for seen and unseen classes, the generator is refined using a supplementary constraint: mapping the generated exemplars to their corresponding class vector.

\paragraph{Transductive ZSL}

All the aforementioned methods suffer from the domain-shift problem, which Transductive Zero-Shot Learning (T-ZSL) approaches \cite{DBLP:journals/corr/abs-1901-01570} %\cite{DBLP:conf/nips/ZhouBLWS03,DBLP:journals/corr/abs-1901-01570} 
attempts to solve.
In T-ZSL, the unlabeled images corresponding to unseen classes are available during training \cite{DBLP:conf/nips/ZhaoDGLXW18}. %\cite{DBLP:conf/pkdd/VermaR17,DBLP:conf/cvpr/YeG17,DBLP:conf/cvpr/SongSYLS18,DBLP:journals/corr/abs-1901-01570}
A wide variety of T-ZSL approaches have been proposed e.g. label propagation \cite{DBLP:conf/icml/FujiwaraI14} 
by exploiting an affinity matrix that measures semantic distances between all classes \cite{DBLP:conf/cvpr/YeG17}, K-means clustering of the visual embedding space \cite{DBLP:journals/corr/abs-1901-01570}, or by projecting unlabeled images to the closest unseen class representation as in DIPL \cite{DBLP:conf/nips/ZhaoDGLXW18}.
%Some methods use label propagation \cite{DBLP:conf/icml/FujiwaraI14}; for example, \cite{DBLP:conf/cvpr/YeG17} "rectify" a prediction matrix obtained by traditional ZSL (matrix of cosine similarities between all images and all classes) to cope with the domain-shift using an affinity matrix that measures semantic distances between all classes.
%Other methods such as DIPL \cite{DBLP:conf/nips/ZhaoDGLXW18} project visual features corresponding to unseen classes close to the closest unseen class representation with a min-min optimization problem. 
%DIPL \cite{DBLP:conf/nips/ZhaoDGLXW18}, a projection matrix is learned to solve an optimization problem that (a) brings projection of visual feature close to semantic representations for seen classes, (b) brings projection of unseen visual features close to the closest unseen class representation. 
%Another set of methods exploit the natural clusters in the visual embedding space, such as  \cite{DBLP:journals/corr/abs-1901-01570} who use a K-means clustering algorithm to determine the centroids of images corresponding to unseen classes. 
However, current models assume the visual space to be well-clustered, which is not the case when the number of classes is high (see \autoref{expes}). 
We demonstrate that such methods fail to benefit from the transductive setting in large-scale settings, by re-implementing one of the best-performing non-generative T-ZSL models: DIPL \cite{DBLP:conf/nips/ZhaoDGLXW18}, and comparing it with our model on ImageNet (20K unseen classes).

\paragraph{Generative models in ZSL and T-ZSL}
\label{gans_in_zsl}

Another body of work \cite{DBLP:conf/iccvw/JurieBH17,DBLP:conf/nips/NiZ019,DBLP:conf/cvpr/HuangWYW19} aims at learning the distributions of visual features conditionally to the semantic representations of labels (e.g. Word2Vec representation of the class): the exponential family can be used, as in GFZSL \cite{DBLP:conf/pkdd/VermaR17}, or more simply Gaussian distributions, as in ZSL-ADA \cite{DBLP:journals/corr/abs-1906-03038}. 
%probability distributions written conditionally to label representations \cite{DBLP:conf/aaai/WangPVFZCRC18,DBLP:conf/iccvw/JurieBH17,DBLP:conf/cvpr/MishraRMM18} using generative models such as Variational Auto-Encoders (VAEs) \cite{DBLP:journals/corr/KingmaW13} or Generative Adversarial Networks (GANs) \cite{DBLP:conf/nips/GoodfellowPMXWOCB14}.
Generative models in (T-)ZSL generally rely on a three-step approach: (i) learn a generator $G$ to produce visual exemplars indistinguishable from the true data distribution (using seen data in inductive models, or from seen and unseen data in transductive models), (ii) generate visual exemplars for unseen classes using $G$, (iii) learn a supervised classifier on these generated data. 
The generator $G$ can be learned using various methods: for example: GANs, as in Gvf \cite{DBLP:conf/iccvw/JurieBH17}, VAEs \cite{DBLP:conf/cvpr/MishraRMM18}, or even CycleGAN, as in ZSL-ADA \cite{DBLP:journals/corr/abs-1906-03038}\footnote{Please note that, despite relying on CycleGAN, our approach is different from ZSL-ADA. Indeed, ZSL-ADA use CycleGAN to learn mappings between generated visual samples and actual visual features, where our CycleGAN model is performed between class labels representations and visual features.}. 
Our approach is not generative, because we use adversarial learning to learn mappings between the distribution of class representations (textual space) and visual features (visual space); these mappings can be used on-the-fly to perform ZSL. 
We show that our model performs better than generative models, more precisely Gvf \cite{DBLP:conf/iccvw/JurieBH17} --- an inductive model, state-of-the-art on ImageNet --- and f-VAEWGAN-D2 \cite{DBLP:conf/cvpr/XianSSA19} ---  a transductive model, state-of-the-art on attribute-based datasets. 

\paragraph{Unsupervised translation using adversarial and cycle-consistency losses}

Adversarial learning aims at estimating a mapping between two data distributions, from non-aligned data. 
In word-to-word translation, \cite{DBLP:conf/iclr/LampleCRDJ18} learn to align two unpaired word spaces from different languages using an adversarial loss.
In image-to-image translation, CycleGAN \cite{cyclegan} has been widely adopted \cite{DBLP:journals/eaai/ZhaoZLHGFJH19};  %\cite{DBLP:journals/corr/LuTT17,DBLP:conf/icml/AlmahairiRSBC18,DBLP:journals/eaai/ZhaoZLHGFJH19}; 
it adds to the adversarial objectives a cycle-consistency objective to constrain mappings to be somewhat invertible. 
%CycleGAN \cite{cyclegan} aims at performing unsupervised translation between two data distributions. More precisely, for two \textit{unpaired} domains $X$ and $Y$, CycleGAN learns two mappings $F:Y\rightarrow X$ and $G:X\rightarrow Y$ by ensuring three constraints: (i) $F(Y) \approx X$ --- via an adversarial loss in $X$, (ii) $G(X) \approx Y$ --- via an adversarial loss in $Y$, (iii) $F(G(X))\approx X$ (and vice versa) --- constraining the mapping to be cycle-consistent.
Despite being widely used with uni-modal data,  %\cite{DBLP:journals/corr/LuTT17,DBLP:conf/icml/AlmahairiRSBC18,DBLP:journals/eaai/ZhaoZLHGFJH19}
CycleGAN has rarely been applied to cross-modal translation, with some exceptions such as speech-to-text alignment \cite{DBLP:conf/nips/ChungWTG18}. However, unlike speech and text, visual and text representations spaces are substantially different, due to the intrinsic semantic discrepancy between language and vision \cite{bruni2014}. To cope with this problem, our model represents images with CONSE embeddings. %We also address the cross-modal \textit{image-sentence matching} retrieval task in a fully unsupervised setting, which is new in the literature. 
%Indeed, some previous works have used adversarial losses (often in \textit{image-sentence matching}), but always as a complement of supervised losses. --- either as a refinement once a standard triplet max-margin loss has been learned \cite{DBLP:journals/tomccap/LiuZWZY19}, or since the beginning of training \cite{DBLP:conf/icmcs/0001XLYSS17,DBLP:conf/mm/WangYXHS17}. 

\begin{figure*}\centering\includegraphics[width = \textwidth]{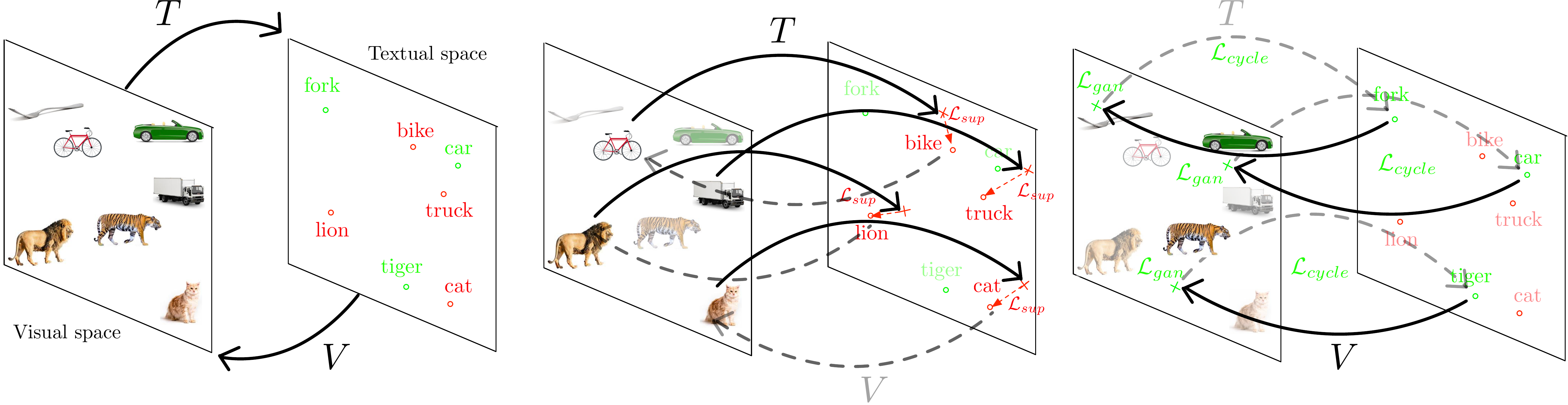}\caption{From right to left: (i) Model overview; our goal is to learn two mapping functions $T$ and $V$ between a visual space $\mathcal{V}$ and a textual space $\mathcal{T}$, where class labels are either \textit{seen} (in red) or \textit{unseen} (in green). To do so, we introduce two objectives: (ii) the \textit{supervised objective} learns to projects images (resp. class labels) to their corresponding class labels (resp. images) for seen classes, and (iii) the \textit{unsupervised objective} is a CycleGAN objective used on the unseen classes to align the visual and textual distributions.}\label{model_overview}\end{figure*}

%\section{Model}
\section{The Cross-Modal CycleGAN Model}

In this section, we present our CM-GAN model for Transductive Zero-Shot Learning, % based on CycleGAN to align unseen class labels with their corresponding images, along with a supervised loss trained on seen classes. 
illustrated in \autoref{model_overview}. Images are embedded in a visual space $\mathcal{V}$, and class labels are embedded in a textual space $\mathcal{T}$. 
%We note $\mathcal{S}$ (resp. $\mathcal{U}$) the set of aligned image-label pairs, and $\mathcal{S}_{\mathcal{T}}$ and $\mathcal{S}_{\mathcal{V}}$ (resp. $\mathcal{U}_{\mathcal{T}}$ and $\mathcal{U}_{\mathcal{V}}$) the distributions of textual and visual vectors from seen classes (resp. unseen classes).  
%We suppose that $\mathcal{V}$ and $\mathcal{T}$ have the same dimension $d$. 
Our goal is to learn cross-modal functions $T$ and $V$ such that $T$ projects images to their corresponding class labels and $V$ projects class labels to their corresponding images. 
Once such functions have been learned, retrieval is made indifferently in $\mathcal{V}$ or $\mathcal{T}$ using a nearest neighbor search. 
Due to the ZSL setting, labels either belong to the seen classes $\mathcal{S}_{\mathcal{T}}$ (marked in red in \autoref{model_overview}) or unseen classes $\mathcal{U}_{\mathcal{T}}$ (green). We note $\mathcal{S}_{\mathcal{V}}$ and $\mathcal{U}_{\mathcal{V}}$ their corresponding images. 
%$\mathcal{U}_{\mathcal{T}}$
% We note $\mathcal{S}_{\mathcal{V}}$ and $\mathcal{U}_{\mathcal{V}}$ the distribution of their corresponding images; $\mathcal{S}$ and $\mathcal{T}$ are the set of aligned image-label pairs. 
% In $\mathcal{S}$, the correspondence between images and labels is known. In $\mathcal{U}$, the correspondence is unknown, but images and labels are accessible during training (\textit{transductive} setting).
%Let us call $\mathcal{S}_t$ (resp. $\mathcal{U}_t$) the set of seen (resp. unseen) classes embeddings and $\mathcal{S}_v$ (resp. $\mathcal{S}_t$) their respective image embeddings
%Following this distinction, we choose to address the transductive ZSL problem using two kinds of information: the \textit{seen information} (section \ref{sourceinf}) and the \textit{unseen information} (section \ref{targetinf}).

%Text-to-Image generation in \cite{DBLP:journals/corr/abs-1808-04538},  %--- following the use of conditional GANs on this task \cite{DBLP:conf/icml/ReedAYLSL16, DBLP:journals/corr/DashGALA17,DBLP:conf/iccv/ZhangXL17,DBLP:journals/pami/ZhangXLZWHM19}

\subsection{Data Representations}
\label{conse}

The first issue that needs to be addressed is to embed modalities so that textual and visual distributions %$\mathbb{P}(\mathcal{V})$ and $\mathbb{P}(\mathcal{T})$ 
are somewhat close and admit a meaningful mapping, as in \cite{DBLP:conf/iclr/LampleCRDJ18} where two unpaired Word2Vec spaces from distinct languages are aligned using adversarial losses. %Their experiments show that the more similar the languages, the better the performance, which suggests that structure of both spaces should be close for the adversarial objective to capture useful information. 
We present below our choices for text and image representations. 

\paragraph{Class labels representation ($T_0$)} Classes are represented with the Skip-Gram embedding of their label \cite{DBLP:conf/nips/MikolovSCCD13}. % -- other embeddings could be used, but in pilot experiments no noticeable change were observed.
%To represent class labels, we use the Skip-Gram algorithm \cite{DBLP:conf/nips/MikolovSCCD13} trained on Wikipedia.
We call $T_0$ the Word2Vec function, trained on Wikipedia, that takes a word as input and outputs a vector of $\mathbb{R}^d$, with $d=500$. If the textual space contains sentences $S$ instead of words, they are represented by the sum of their word embeddings: $s=\sum_{w \in S} T_0(w)$.

%; for images, we use the CONSE model \cite{DBLP:journals/corr/NorouziMBSSFCD13}, where images are embedded using a linear combination of word vectors (the choice of CONSE is further explained in subsection \ref{conse}). 
\paragraph{Image representation ($V_0$)} We call $V_0$ the image embedding function, which transforms an input image into a vector of $\mathbb{R}^d$. 
 %for example \cite{cyclegan} tackle image-to-image unsupervised translation, and \cite{DBLP:conf/iclr/LampleCRDJ18} tackle word-to-word unsupervised translation. 
We argue that visual representations should be ``homogeneous'' to the classes embeddings so that distributions can be mapped on one another. Thus, we propose to use the CONSE model \cite{DBLP:journals/corr/NorouziMBSSFCD13}, in which an image is represented as a convex combination of class label embeddings. %, where the weight is linked to the predicted class probability. 
More precisely, for a given image, let us call $p(i)$ the distribution over the seen classes ${i \in [1,|\mathcal{S}_{\mathcal{T}}|]}$ output by the CNN, and $T_K$ the indices of the $K$ most probable classes. The representation of an image is then:

\begin{equation}
V_0(v)=   \sum_{i \in T_K}p(i | i \in T_K) T_0(l_i) 
\end{equation}{}

\noindent where $l_i$ is the label of class $i$ and $p(i | i \in T_K)= \frac{p(i)}{\sum_{j \in T_K}p(j)}$. 
%: we define $(\beta_i)_{i \in [1,N]}$ as the normalization of $(\delta_{i \in T_K}. \alpha_i)_{i \in [1,N]}$ so that $\sum_{i=1}^N \beta_i=1$. The visual representation is defined as: $\sum_{i \in [1,K]}\beta_i w_i$  where $w_i$ is the word representation of class $i$. 

To assess our intuition that CONSE leads to a visual space that is semantically close to the textual (Word2Vec) modality, we conduct a preliminary experiment by comparing CONSE to other visual features: (a) vectors randomly sampled from a normal distribution (Random), (b) CNN \cite{DBLP:conf/cvpr/HeZRS16} and (c) DeViSE \cite{devise} embeddings; results are reported in \autoref{mnno}. 
To do so, we use the $\rho_{vis}$ metric \cite{DBLP:conf/emnlp/BordesZSPG19}, which measures the similarity of two sets of vectors even if they do not share a joint embedding space. %More precisely, $\rho_{vis}$ measures the correlation between the cosine similarity of two class labels in the textual space with the cosine similarity of their corresponding images in the visual space. 
More precisely, we set $\rho_{vis}(\mathcal{S})=\rho(cos(t,t'),cos(v,v'))$ where $\rho$ is the Pearson correlation and pairs $v,t$ and $v',t'$ are aligned and sampled from $\mathcal{S}$; similarly, $\rho_{vis}(\mathcal{U})$ can be defined when pairs are sampled from unseen data. % or $\mathcal{U}$. %Thus, this metric can measure  %in  \cite{DBLP:conf/acl/CollellM18} 
%as the proportion mean cosine similarity of $K$ nearest neighbors shared by paired elements of both modalities measured in $\%$ ($K=10$). 
%Thus, this metric measures the similarity of two sets of vectors. 

We observe that DeViSE features obtain higher $\rho_{vis}$ scores compared to CNN features (53.8 vs 30.7, 36.9 vs 29.4): this shows that learning a projection between the visual and the textual space (DeViSE) leads to more meaningful representations compared to CNN features, where the semantics of class labels is not used. However, being based on class labels representations, CONSE features show an even higher similarity between modalities compared to DeViSE (92.1 vs 53.8, 45.9 vs 36.9). This experiment confirms preliminary results, where adversarial losses failed to produce meaningful models when applied to CNN or DeViSE vectors.  % This confirms our choice, and gives a hint of why optimizing the CycleGAN objective leads to a better cross-modal alignment in the case of CONSE, whereas the CycleGAN objective crashes instantaneously when applied to CNN or DeViSE embeddings. 

Having set the initial representations of class labels and images, we now proceed to define the training losses (\autoref{sourceinf} and \autoref{targetinf}) and the learning procedure (\autoref{learning}).

\begin{table}[t]                            \centering 
\begin{tabular}{l c c }
\toprule
Model  &  $\mathcal{S}$ & $\mathcal{U}$\\
\hline 
Random & 11.6& 11.6\\ %1.\\%2.79 for 360 \\ 
CNN  \cite{DBLP:conf/cvpr/HeZRS16} &  30.7& 29.4\\ %& 15.1 \\%& 2.3  \\
DeViSE \cite{devise}& 53.8 & 36.9\\ %23.5  \\
CONSE  \cite{DBLP:journals/corr/NorouziMBSSFCD13}& \textbf{92.1} & \textbf{45.9}\\ %58.2 \\ %38.09 for 360 \\
\bottomrule
\end{tabular} 
\vspace{-0.2cm}
\caption{Preliminary experiment: comparison of visual representations. The metric is $\rho_{vis}$ (in $\%$) computed on ImageNet's  $\mathcal{S}$ (1K seen classes) and $\mathcal{U}$ (20K unseen classes) --- the higher the better. 
CONSE and DeViSE were re-implemented.  %dataset (360 unseen classes, 400K images).
}
\label{mnno}
%\vspace{-0.3cm}
\end{table}

\subsection{Supervised Loss}
\label{sourceinf}

%N1\begin{multline} \mathcal{L}_{sup}= \hspace{-0.4cm} \mathop{\mathbb{E}}\limits_{v,t,v^{-},t^{-}} % \in T_{\mathcal{S}}^{2} * V_{\mathcal{S}}^2}%\hspace{0.2cm}(\big\lfloor \gamma - \cos(C_T(v), M_T(t))  + \cos(C_T(v), M_T(t^-)) \big\rfloor_{+} \\
%\big\lfloor  \gamma - \cos(M_V(v), C_V(t))  + \cos(M_V(v^-), M_V(t)) \big\rfloor_{+})
%\end{multline}

%N1
%\begin{multline}
% \mathcal{L}_{sup}= \hspace{-0.4cm} \mathop{\mathbb{E}}\limits_{v,t,v^{-},t^{-}} % \in T_{\mathcal{S}}^{2} * V_{\mathcal{S}}^2}%\hspace{0.2cm}
%(\big\lfloor \gamma - \cos(T(v), T(t))  + \cos(T(v), T(t^-)) \big\rfloor_{+} \\
%\big\lfloor  \gamma - \cos(V(v), V(t))  + \cos(V(v^-), V(t)) \big\rfloor_{+})
%\end{multline}

%N2
%\begin{multline}
% \mathcal{L}_{sup}= \hspace{-0.4cm} \mathop{\mathbb{E}}\limits_{v,t,v^{-},t^{-}} % \in T_{\mathcal{S}}^{2} * V_{\mathcal{S}}^2}%\hspace{0.2cm}
%(\big\lfloor \gamma - \cos(v_T, t_T)  + \cos(v_T, t^-_T) \big\rfloor_{+} \\
%\big\lfloor  \gamma - \cos(v_T, t_T)  + \cos(v^-_T, t_T) \big\rfloor_{+})
%\end{multline}

%
%\begin{multline}
% \mathcal{L}_{sup}= \hspace{-0.4cm} \mathop{\mathbb{E}}\limits_{v,t,v^{-},t^{-}} % \in T_{\mathcal{S}}^{2} * V_{\mathcal{S}}^2}%\hspace{0.2cm}
%(\big\lfloor \gamma - \cos(T_v, T_t)  + \cos(T_v, T_{t^-}) \big\rfloor_{+} \\
%\big\lfloor  \gamma - \cos(T_v, T_t)  + \cos(T_{v^-}, T_t) \big\rfloor_{+})
%\end{multline}

The \textit{supervised loss} leverages the information of seen classes, as illustrated in \autoref{model_overview} (middle). 
The correspondence between images and seen class labels is a many-to-one correspondence, that can be exploited using a standard max-margin triplet loss $\mathcal{L}_{sup}$. Indeed, triplet losses are commonly used \cite{devise} to bring closer elements from distinct modalities in a common space, and have been shown \cite{DBLP:journals/jmlr/WeinbergerS09} to be more efficient than alternatives such as hinge-loss functions \cite{DBLP:conf/cvpr/HadsellCL06} to learn meaningful cross-modal projections. The supervised loss $\mathcal{L}_{sup}$ is defined by:
%\vspace{-0.1cm}
\begin{multline}
 \mathcal{L}_{sup}= \hspace{-0.2cm} \mathop{\mathbb{E}}\limits_{v,t,v^-,t^-} %\in \mathcal{S}*\mathcal{S}_{\mathcal{V}}*\mathcal{S}_{\mathcal{V}}} % \in T_{\mathcal{S}}^{2} * V_{\mathcal{S}}^2}%
%  \hspace{-0.8cm}
(\big\lfloor \gamma - \cos(T(\hat v), \hat t)  + \cos(T(\hat v), {\hat t}^-) \big\rfloor_{+} \\
+ \big\lfloor  \gamma - \cos(\hat v, V(\hat t))  + \cos({\hat v}^-, V(\hat t)) \big\rfloor_{+})
\label{suploss}
\end{multline}

\noindent where $v$ is an image with $t$ its label sampled from the seen images, and $v^{-}$ and $t^{-}$ are negative examples sampled from the set of images and labels respectively. 
%from $\mathcal{S}_{\mathcal{V}}$ and  $\mathcal{S}_{\mathcal{T}}$. 
We denote $\hat v$ (resp.\ $\hat t$) the current representation of the image $v$ (resp.\ of label $t$) %(and similarly for $t$, $v^-$ and $t^-$) 
that we define precisely in \autoref{learning}. 
The margin is used to tighten the constraint: the discrepancy between cosine similarities for positive and negative pairs should be higher than a fixed margin $\gamma$, for which we use a standard value: $\gamma=0.5$. 
%We provide more details on $T$ and $V$ in Section \ref{learning}.
%where $T$ and $V$ are the distributions of textual and visual vectors in their respective spaces 
%\begin{align}
%\mathcal{L}_{sup} &= \mathop{\mathbb{E}}\limits_{i, v \sim \mathcal{S}} \mathop{\mathbb{E}}\limits_{j \sim \mathcal{S}}
%\big\lfloor \gamma - cos(F(v), w_i)  + cos(F(v), w_j) \big\rfloor_{+}%  +  \big\lfloor \gamma - cos(v, G(w_i))  + cos(v, G(w_j)) \big\rfloor_{+}) \label{L_V} \notag
%\end{align}     

%\begin{figure} \centering \includegraphics[width = \columnwidth]{lsup.pdf} \caption{seen information} \label{source_information} \end{figure}

\subsection{Adversarial and Cycle-Consistency Losses}
\label{targetinf}

The \textit{transductive loss} aims at capturing information from the unseen classes, as illustrated on the right of \autoref{model_overview}. 
Since there is no known mapping between unseen class labels and their corresponding images, we can only use the distribution of images and classes to align modalities. We use CycleGan \cite{cyclegan}, which has proven useful in the unpaired image-to-image translation task. The learned mappings are the cross-modal functions $T$ and $V$, and discriminators in each modality $D_{\mathcal{V}}$ and $D_{\mathcal{T}}$. The CycleGan objective consists of adversarial losses in both spaces $\mathcal{L}_{gan}^v$ and $\mathcal{L}_{gan}^t$ combined to a cycle-consistency loss $\mathcal{L}_{c}$ weighted by a scalar $\lambda_c$: 

%$$\mathcal{L}_{cgan} =\underbrace{\mathcal{L}_{GAN}^v + \mathcal{L}_{GAN}^t}_{\mathcal{L}_{gan}} + \lambda_c.\mathcal{L}_{cycle} $$

%\vspace{-0.2cm}
\begin{equation}
%\hspace{-0.3cm} \mathcal{L}_{cgan} =\mathcal{L}_{gan}^v(T,D_{\mathcal{T}}) + \mathcal{L}_{gan}^t(V,D_{\mathcal{V}}) + \lambda_c.\mathcal{L}_{c}(V,T) 
\mathcal{L}_{cgan} =\mathcal{L}_{gan}^v + \mathcal{L}_{gan}^t + \lambda_c.\mathcal{L}_{c}
\label{unsup}
\end{equation}
%\vspace{-0.3cm}

%$$\mathcal{L}_{cgan} =\mathcal{L}_{gan}^v(T,D_{\mathcal{T}}) + \mathcal{L}_{gan}^t(V,D_{\mathcal{V}}) + \lambda_c.\mathcal{L}_{c}(F,G) $$

\noindent The individual losses write as follows:
%\vspace{-0.3cm}

\begin{equation}
\mathcal{L}_{gan}^v = \mathop{\mathbb{E}}\limits_{t \in \mathcal{U}_{\mathcal{T}}}[\log D_{\mathcal{T}}(\hat t)] +  \mathop{\mathbb{E}}\limits_{v \in \mathcal{U}_{\mathcal{V}}}[\log(1-D_{\mathcal{T}}(T(\hat v)))]
\end{equation}
%\vspace{-0.45cm}
\begin{equation}
\mathcal{L}_{gan}^t = \mathop{\mathbb{E}}\limits_{v \in \mathcal{U}_{\mathcal{V}}}[\log D_{\mathcal{V}}(\hat v)] +  \mathop{\mathbb{E}}\limits_{t \in \mathcal{U}_{\mathcal{T}}}[\log(1-D_{\mathcal{V}}(V(\hat t)))]
\end{equation}
%\vspace{-0.3cm}
\begin{equation}
\mathcal{L}_{c} = \mathop{\mathbb{E}}\limits_{t \in \mathcal{U}_{\mathcal{T}}}[\|T(V(\hat t)) - \hat t \|_2] + \mathop{\mathbb{E}}\limits_{v \in \mathcal{U}_{\mathcal{V}}}[\|V(T(\hat v)) - \hat v \|_2]
\end{equation}
%\vspace{-0.25cm}

%where $T$ and $V$ are the distributions of textual and visual vectors in their respective spaces (see Section \ref{learning} for more detail).  
%$$\mathcal{L}_{gan}^v = \mathbb{E}_{t \in \mathcal{T}}[\log D_{\mathcal{T}}(t)] +  \mathbb{E}_{v \in \mathcal{V}}[\log(1-D_{\mathcal{T}}(T(v)))]$$
%$$\mathcal{L}_{gan}^t = \mathbb{E}_{v \in \mathcal{V}}[\log D_{\mathcal{V}}(v)] +  \mathbb{E}_{t \in \mathcal{T}}[\log(1-D_{\mathcal{V}}(V(t)))]$$
%$$\mathcal{L}_{c} = \mathbb{E}_{t \in \mathcal{T}}[\|T(V(t)) - t \|_2] + \mathbb{E}_{v \in \mathcal{V}}[\|V(T(v)) - v \|_2]$$

\noindent where  $\mathcal{U}_{\mathcal{T}}$ and $\mathcal{U}_{\mathcal{V}}$ are the textual and visual distributions for unseen classes.  
$V$ (resp. $T$) aims at generating visual representations that are indistinguishable from vectors of $\mathcal{U}_{\mathcal{V}}$ (resp. $\mathcal{U}_{\mathcal{T}}$), whereas the discriminator $D_{\mathcal{V}}$ (resp. $D_{\mathcal{T}}$) aims at distinguishing elements from $\mathcal{U}_{\mathcal{V}}$ (resp. $\mathcal{U}_{\mathcal{T}}$) and elements from $V(\mathcal{U}_{\mathcal{T}})$ (resp. $T(\mathcal{U}_{\mathcal{V}})$). As in \cite{DBLP:conf/nips/GoodfellowPMXWOCB14}, $V$ (resp. $T$) aims at minimizing $\mathcal{L}_{GAN}^t$ (resp. $\mathcal{L}_{GAN}^v$), and the discriminator $D_{\mathcal{V}}$ (resp. $D_{\mathcal{T}}$) aims at maximizing it in an adversarial fashion. For the cycle-consistency loss $\mathcal{L}_c$, using the L2 norm led to better results than the L1 norm, which is used in the original CycleGAN model.
%There is more data in the visual space: better to train the discriminator/generator in the visual space (thus, better to learn $V$)

%\begin{figure} \centering \includegraphics[width=\columnwidth]{cgan.pdf} \caption{unseen information} \label{target_information} \end{figure}

\subsection{Learning}
\label{learning}

To learn the cross-modal functions $T$ and $V$, we adopt an alternate iterative learning procedure, as preliminary experiments showed that jointly optimizing $\mathcal{L}_{cgan}$ and $\mathcal{L}_{sup}$ leads to highly unstable training. 
Instead of learning $T$ and $V$ in one step, we refine these mappings iteratively by composing the functions $T_k$ and $V_k$ learned at each optimization step $k$, giving the global mappings $\hat{T}_k$ and $\hat{V}_k$. Thus, the supervised and transductive losses  (equations \eqref{suploss} and \eqref{unsup}) are optimized using data from the previous step: $\hat v = \hat{T}_{k-1}(v)$ for each image $v$ and $\hat t = \hat{V}_{k-1}(t)$ for each class label $t$; these representations are fixed to avoid over-fitting. 
%Instead of working with fixed representation of images and text for all iteration steps, we learn a succession of projection functions, and start each step with the projected image and text vectors from the previous step. 

%Let us denote $\hat{T}_k$ (resp. $\hat{V}_k$) the projection from the image (resp. text) space to the visual (resp. textual) semantic space.
%At each optimization step $k$, we use the representation $\hat v = \hat{T}_{k-1}(v)$ for each image $v$ and $\hat t = \hat{V}_{k-1}(t)$ for each text $t$ in the equations \eqref{suploss} and \eqref{unsup}.

% $\mathcal{S}_0 = \{(V_0(v), T_0(t)) | (v,t) \in \mathcal S\}$ where $\mathcal S$ is the set of image with their corresponding label, and similarly for the unseen set $\mathcal{U}_0$.

\paragraph{Supervised step}

During a supervised step $k$, we optimize $T_k$ and $V_k$ -- modeled as 2-layer Perceptrons -- with equation \eqref{suploss}.
We notice experimentally that our validation measure is optimal when retrieval is performed in the textual space $\mathcal{T}$ (cf. \autoref{implem}), probably due to the many-to-one mapping that provides $T_k$ with significantly more training data than $V_k$ as input (1.3M images versus 1K seen classes). Thus, we use the learned $T_{k}$ to update data representations, and we set $\hat{T}_k = T_k \circ \hat{T}_{k-1}$ and $\hat{V}_k = \hat{V}_{k-1}$. 

\paragraph{Transductive step}

During a transductive step $k$, we optimize $T_k$ and $V_k$ -- modeled as 2-layer Perceptrons -- with equation \eqref{unsup}.
We notice experimentally that our validation measure is optimal when retrieval is performed in the visual space $\mathcal{V}$, probably because the discriminator $D_{\mathcal{V}}$ has significantly more training data on which to be trained (13M images versus 20K unseen classes), thus leading to a better cross-modal $V_k$.
Consequently, we use the learned $V_k$, and we set $\hat{T}_k =  \hat{T}_{k-1}$ and $\hat{V}_k = V_k \circ \hat{V}_{k-1}$.

\paragraph{Outcome}

When the validation measure (cf. \autoref{implem}) shows no improvement between step $K$ and step $K+1$, training is stopped, and we obtain the final functions $\hat{V}_K$ and $\hat{T}_K$.

\section{Experimental Protocol}

\subsection{Datasets}

\paragraph{ImageNet} ImageNet \cite{imagenet} consists of 14.2M images, corresponding to 21841 classes, 1000 of which are seen classes, and the rest unseen. Among all unseen classes, we keep 20345 classes that have a Word2Vec embedding (compound words are averaged); we used the same word embeddings and same classes as \cite{DBLP:conf/cvpr/ChangpinyoCGS16}. We note \textit{ImageNet-Full} the dataset with all 20345 unseen classes. 
Among these classes, \textit{2-hop} (resp.\ \textit{3-hop}) consists of 1509 (resp.\ 7678) classes within two (resp.\ three) hops of a seen class in WordNet. Thus, \textit{2-hop} classes are semantically close to the training classes, and \textit{3-hop} classes are comparatively farther (thus leading to better performances on \textit{2-hop} than \textit{3-hop}); evaluating a model on these sets is interesting as it allows to measure the model's capacity to generalize its learned knowledge. We also consider \textit{ImageNet-360}, which is widely adopted among the ZSL literature because it is substantially smaller and thus allows comparison with the literature (360 unseen classes, with 400K images). 
 
\paragraph{MS COCO} We use the MS COCO dataset \cite{DBLP:conf/eccv/LinMBHPRDZ14} to tackle a new task that we introduce in this paper, that we call \textit{zero-shot sentence-to-image matching}. We suppose that no text-to-image correspondence is known, and we evaluate our model on cross-modal retrieval. This task is interesting as (i) no supervised information can be used, (ii) it features sentences instead of words, and (iii) it extends ZSL to a very high number of classes (as many classes as sentences). 
The training set consists of 118K images, with 5 captions per image. Evaluation is performed over 1K images (along with the corresponding 5K captions) from the test set of MS COCO. %In this experiment, there is no aligned data, no correspondence between images and sentences: thus, there is no the unsupervised information of unseen classes to be exploited. 

\subsection{Evaluation Metrics}

In the experiments, we consider the two standard evaluation settings: \textbf{Zero-Shot Learning} (ZSL), in which the image label is searched among unseen classes $\mathcal{U}$, and \textbf{Generalized Zero-Shot Learning} (G-ZSL), where the class is searched among seen and unseen classes. 

Following \cite{DBLP:conf/nips/ZhaoDGLXW18} %\cite{DBLP:conf/cvpr/ChangpinyoCGS16,DBLP:conf/iccv/ChangpinyoCS17,DBLP:conf/nips/ZhaoDGLXW18}
, we use the Flat-Hit@$k$ $\in \{1,2,5,10,20\} $ metric (noted $\text{FH}_k$), which is the metric used in the vast majority ZSL literature applied to ImageNet. 
$\text{FH}_k$ is defined as the percentage of images for which the correct label is present in the top $k$ predictions of the model. % (a.k.a.\ Recall@$k$ metric). 

Furthermore, following \cite{DBLP:conf/icml/ZablockiBSPG19}, we use the Mean First Relevant (noted MFR) metric to evaluate our model scenarios, as this metric is more stable than $\text{FH}_k$ and is not sensitive to the number of classes, thus enabling fine-grained model comparisons.
MFR is defined as the mean value of the rank of the correct class (noted FR) among the model's predictions, averaged over the set of test images $\mathcal{U}_{\mathcal{V}}$ and linearly re-scaled %with a factor $\frac{100}{|N_{im}^{test}|}$ 
so that the random model has a $50 \%$ score: 

%\vspace{-0.3cm}
\begin{equation}
    \text{MFR} = \frac{100}{K |\mathcal{U}_{\mathcal{V}}| } \sum_{v \in \mathcal{U}_{\mathcal{V}}} \text{FR}_v 
\end{equation}{}

\noindent where $K=|\mathcal{U}_{\mathcal{T}}|$ for ZSL and $K=|\mathcal{U}_{\mathcal{T}}| + |\mathcal{S}_{\mathcal{T}}|$ for G-ZSL.

\subsection{Baselines}

For ImageNet, we compare CM-GAN to a variety of approaches:

\begin{itemize}
    \item Standard (inductive) ZSL: \textbf{DeViSE} \cite{devise}, \textbf{CONSE} \cite{DBLP:journals/corr/NorouziMBSSFCD13}, \textbf{SYNC} \cite{DBLP:conf/cvpr/ChangpinyoCGS16} and \textbf{EXEM} \cite{DBLP:conf/iccv/ChangpinyoCS17};
    \item Inductive generative: GAN-based model \textbf{Gvf} \cite{DBLP:conf/iccvw/JurieBH17} (state-of-the-art on ImageNet-Full) and  VAE-based models \textbf{VZSL} \cite{DBLP:conf/aaai/WangPVFZCRC18}, \textbf{CVAE-ZSL} \cite{DBLP:conf/cvpr/MishraRMM18}  and \textbf{SE-ZSL}  \cite{DBLP:conf/cvpr/VermaAMR18};
    \item Transductive (non-generative): \textbf{DIPL} \cite{DBLP:conf/nips/ZhaoDGLXW18} (state-of-the-art on ImageNet-360). In DIPL, a projection matrix is learned to solve an optimization problem that (a) brings projection of visual feature close to semantic representations for seen classes, (b) brings projection of unseen visual features close to the closest unseen class representation;
    \item Transductive generative: \textbf{f-VAEWGAN-D2} \cite{DBLP:conf/cvpr/XianSSA19} (state-of-the-art on attribute-based datasets). f-VAEGAN-D2, like most generative models, aims at learning a generator $G$ to produce visual samples for unseen classes, and then train a supervised classifier on these synthetic data. Here, the generator $G$ is trained on three objectives: (i) WGAN objective: $G$ generates visual samples for seen classes, from Gaussian noise concatenated with the class vector, and a discriminator is trained to distinguish between $G$’s outputs and true data, 
(ii) VAE objective: a VAE is trained to reconstruct data from seen classes, and $G$ is the VAE’s decoder,
and (iii) transductive objective: $G$ is the generator, and a discriminator is trained to distinguish between generated sampled for unseen classes and true samples.  

\end{itemize}{}

All reported results are extracted from the original papers, which explains that we could not report all metrics for all models. For DIPL and for f-VAEWGAN-D2, we re-implement their model to get ImageNet-Full scores. We note CONSE* our re-implementation of CONSE, which performs better than the original CONSE \cite{DBLP:journals/corr/NorouziMBSSFCD13} since we use the CNN network Inception-V3 \cite{inception}, which shows better performances than AlexNet \cite{DBLP:journals/cacm/KrizhevskySH17}.

For MS COCO, as our unsupervised setting is new in the literature, we only compare to a \textit{supervised} baseline, for which we suppose that the sentence-to-image alignment is known and 
the $\mathcal{L}_{sup}$ objective is optimized to map CONSE representation to the corresponding sentence vectors. 
%we know the alignment between images and sentences, and we optimize the $\mathcal{L}_{sup}$ objective.  

\subsection{Implementation Details}
\label{implem}
%In all experiments, $T$ and $V$ are 1-layer Perceptrons. %Images are represented using the method explained in Subsection \ref{conse}. 

In the experiments performed on ImageNet-Full and ImageNet-360, our validation metric is the value of the max-margin triplet loss $\mathcal{L}_{sup}$ computed on the seen classes. The choices described in \autoref{learning} (retaining $T_k$ for the supervised step and $V_k$ for the transductive step), were made by computing separately both rows of \autoref{suploss}, to determine in which space retrieval is optimal. %--- indeed, presuming that the unsupervised step has to improve retrieval for source classes seems a fair assumption to assess the effect of $\mathcal{L}_{cgan}$. 
This metric is used to determine $\lambda_c \in \{1,5,10\}$ and the stopping step. Our final model is optimal at $K=6$ steps. %, in the sense that the $(K+1)$-th iteration shows no improvement regarding the  validation criterion. 
Selected parameters for the transductive steps are respectively $\lambda_c=1, 10, 1$.  

In the experiments performed on MS COCO, our validation criterion is the unsupervised criterion described in \cite{DBLP:conf/iclr/LampleCRDJ18}: the mean cosine similarity between a set of images and their predicted sentences (from a selected 1K images/ 5K captions of the validation set). Due to the absence of supervised data, there is no iterative process, but only $K=1$ transductive step where $\mathcal{L}_{cgan}$ is optimized. Thus, as explained in \autoref{learning}, $V_1$ is used to obtain final cross-modal functions: $\hat{T}_K =  T_{0}$ and $\hat{V}_K = V_1 \circ V_{0}$. 

All images were processed with Inception-V3 \cite{inception} to build CONSE embeddings. The code was implemented using PyTorch.  % --- we will note CONSE* our re-implementation, which performs better than the original CONSE \cite{DBLP:journals/corr/NorouziMBSSFCD13} which is based on the lesser performing AlexNet \cite{DBLP:journals/cacm/KrizhevskySH17}. 

%\section{Results}

%In this section, we perform extensive experiments to show the effectiveness of CM-GAN on the ImageNet T-ZSL task (Section \ref{expes}) --- on which a post-analysis on word embeddings is provided (Section \ref{grounding}) and on a task we introduce: \textit{zero-shot sentence-to-image matching} on MS COCO (Section \ref{section_mscoco}).

%First, in section \ref{expes}, we evaluate CM-GAN on the ImageNet ZSL task and provide SOTA results. 
%Then, in section \ref{grounding}, we learn and evaluate grounded word representations using supervised and unsupervised visual information.
%Then, in section \ref{section_mscoco}, we evaluate our Cross-modal CycleGAN model on an unpaired sentence-to-image matching task on MS COCO.

%\section{Results}

\subsection{Zero-Shot Learning on ImageNet}
\label{expes}

\begin{table}[t]                                            
    \centering 
      %  \begin{tabularx}{\columnwidth}{m{5pt}   m{5pt}| c |  p{20pt} p{21pt} p{19pt} p{19pt} p{20pt}  }
      
        \begingroup
        \renewcommand*{\arraystretch}{0.77}
        
        \begin{tabular}{@{}c l ccccc@{}} 
        
        \toprule
        
     & Model & $\text{FH}_{1}$ & $\text{FH}_{2}$ & $\text{FH}_{5}$ & $\text{FH}_{10}$ & $\text{FH}_{20}$\\
    \hline
       \multirow{9}{*}{\rotatebox[origin=c]{90}{All}}  & CONSE \cite{DBLP:journals/corr/NorouziMBSSFCD13}
    &1.4& 2.2 & 3.9 & 5.8 & 8.3  \\  
 & CONSE* \cite{DBLP:journals/corr/NorouziMBSSFCD13} & 1.76 & 2.8 & 4.77& 7.07& 10.04 \\
 & DeViSE \cite{devise}& 0.8 & 1.4 & 2.5 & 3.9 & 6.0\\
     & SYNC \cite{DBLP:conf/cvpr/ChangpinyoCGS16}
     & 1.5& 2.4 & 4.5 & 7.1 & 10.9\\ 
      & DIPL* \cite{DBLP:conf/nips/ZhaoDGLXW18} & 1.47 & 2.52 & 4.82 & 7.59 & 11.52 \\
     & EXEM \cite{DBLP:conf/iccv/ChangpinyoCS17}
     & 1.8 & 2.9 & 5.3 & 8.2 & 12.2\\
 & f-VAEWGAN-D2* \cite{DBLP:conf/cvpr/XianSSA19} & 1.82 &2.95& 5.41& 8.22& 12.45 \\
     & Gvf \cite{DBLP:conf/iccvw/JurieBH17}     & 1.90& 3.03 & 5.67 & 8.31 & \textbf{13.14}\\
     &   \textbf{CM-GAN} & \textbf{1.99} & \textbf{3.18} & \textbf{5.76} & \textbf{8.64} & 12.57\\[0.3cm]
    
    %\hline
    
      \multirow{9}{*}{\rotatebox[origin=c]{90}{$3_{\text{hop}}$}}  & CONSE \cite{DBLP:journals/corr/NorouziMBSSFCD13}
    & 2.7 & 4.4 & 7.8 & 11.5 & 16.1 \\  
 & CONSE* \cite{DBLP:journals/corr/NorouziMBSSFCD13} & 3.43 & 5.22 & 8.96 & 12.96& 18.21 \\
 & DeViSE \cite{devise}  & 1.7 & 2.9 & 5.3& 8.2 & 12.5\\
     & SYNC \cite{DBLP:conf/cvpr/ChangpinyoCGS16}
    & 2.9& 4.9 & 9.2 & 14.2 & 20.9\\ 
      & DIPL* \cite{DBLP:conf/nips/ZhaoDGLXW18}& 2.81 & 5.02 & 9.87 & 15.64 & 22.1 \\
     & EXEM \cite{DBLP:conf/iccv/ChangpinyoCS17}
    & 3.6 & 5.9 & 10.7 & 16.1 & 23.1 \\
     & f-VAEWGAN-D2* \cite{DBLP:conf/cvpr/XianSSA19}& 3.33 & 5.91& 10.83& 16.2& 23.34\\
     & Gvf \cite{DBLP:conf/iccvw/JurieBH17}    & 3.58 & 5.97 & 11.03 & 16.51 & \textbf{23.88}\\
     &   \textbf{CM-GAN} & \textbf{3.88} & \textbf{6.15} & \textbf{11.25} & \textbf{16.66} & 23.4 \\[0.3cm]
   
    %\hline
    \multirow{9}{*}{\rotatebox[origin=c]{90}{$2_{\text{hop}}$}}  & CONSE \cite{DBLP:journals/corr/NorouziMBSSFCD13} 
    & 9.4 & 15.1 & 24.7 & 32.7 & 41.8 \\ 
 & CONSE* \cite{DBLP:journals/corr/NorouziMBSSFCD13}& 10.67 & 15.58 & 25.24 & 34.29 & 45.31\\
 & DeViSE \cite{devise}  & 6.0 & 10.0 &18.1 & 26.4 & 36.4\\
    & SYNC \cite{DBLP:conf/cvpr/ChangpinyoCGS16}
   & 10.5 & 16.7 & 28.6 & 40.1 & 52.0\\ 
      & DIPL* \cite{DBLP:conf/nips/ZhaoDGLXW18}& 10.46 & 16.79 & 28.23& 39.4&52.08 \\
      &EXEM \cite{DBLP:conf/iccv/ChangpinyoCS17}
   & 12.5 & 19.5 & 32.3 & 43.7 & 55.2 \\
    & f-VAEWGAN-D2* \cite{DBLP:conf/cvpr/XianSSA19}& 13.2 & 20.09& 32.87& 43.72& 56.03\\
      & Gvf \cite{DBLP:conf/iccvw/JurieBH17}
   & 13.05 &\textbf{21.52} & 33.71 & 43.91 &\textbf{57.31} \\
      &   \textbf{CM-GAN} & \textbf{13.7}& 20.96 & \textbf{33.73} & \textbf{45.51} & 56.31 \\

    %\cline{2-8}
    \bottomrule
    
    \end{tabular}    
    \endgroup
    
    \caption{Zero-Shot Learning results on ImageNet-Full. Models marked with * were re-implemented.}
    \label{zsl_imagenet}
\end{table}

\begin{table}[t]                                            
    \centering 
      %  \begin{tabularx}{\columnwidth}{m{5pt}   m{5pt}| c |  p{20pt} p{21pt} p{19pt} p{19pt} p{20pt}  }
      
        \begingroup
        \renewcommand*{\arraystretch}{0.77}
        
        \begin{tabular}{@{}c l ccccc@{}} 
        
        \toprule 
        
     & Model & $\text{FH}_{1}$ & $\text{FH}_{2}$ & $\text{FH}_{5}$ & $\text{FH}_{10}$ & $\text{FH}_{20}$\\

\hline

  \multirow{6}{*}{\rotatebox[origin=c]{90}{All }}  & CONSE \cite{DBLP:journals/corr/NorouziMBSSFCD13}
   & 0.2 & 1.2 & 3.0 & 5.0 & 7.5  \\  
     & CONSE* \cite{DBLP:journals/corr/NorouziMBSSFCD13} & 0.13 & 1.41 & 3.62 & 5.97 & 8.93 \\
     & DeViSE \cite{devise}& 0.3 & 0.8 & 1.9 & 3.2 & 5.3\\
     & f-VAEWGAN* \cite{DBLP:conf/cvpr/XianSSA19}& 0.89& 1.82& 4.31 & 6.05&9.8 \\
     & Gvf \cite{DBLP:conf/iccvw/JurieBH17}
     & \textbf{1.03} & \textbf{1.93} & \textbf{4.98} & 6.23 & 10.26\\
     &   \textbf{CM-GAN} & 0.16 & 1.5 & 4.24 & \textbf{7.28} & \textbf{11.28} \\[0.3cm]
     
   \multirow{6}{*}{\rotatebox[origin=c]{90}{$3_{\text{hop}}$ }}  & CONSE \cite{DBLP:journals/corr/NorouziMBSSFCD13}
   & 0.2 & 2.4 & 5.9 & 9.7 & 14.3 \\ 
    & CONSE* \cite{DBLP:journals/corr/NorouziMBSSFCD13}& 0.21 & 2.65 & 6.76 & 10.77& 16.01 \\
   & DeViSE \cite{devise} & 0.5 & 1.4 & 3.4 & 5.9 & 9.7 \\
 & f-VAEWGAN-D2* \cite{DBLP:conf/cvpr/XianSSA19}& 1.85& 3.88& 6.42& 10.94& 16.21\\
    & Gvf \cite{DBLP:conf/iccvw/JurieBH17}
    & \textbf{1.99} & \textbf{4.01} & 6.74 & 11.72 & 16.34\\
     &   \textbf{CM-GAN} & 0.26 & 2.65 & \textbf{7.94} & \textbf{13.73} & \textbf{20.56} \\[0.3cm]
   
    \multirow{6}{*}{\rotatebox[origin=c]{90}{$2_{\text{hop}}$ }}  & CONSE \cite{DBLP:journals/corr/NorouziMBSSFCD13}
    & 0.3 & 7.1 & 17.2 & 24.9 & 33.5 \\  
 & CONSE* \cite{DBLP:journals/corr/NorouziMBSSFCD13}& 0.17 & 6.3 & 16.37 & 24.67 & 34.45 \\
 & DeViSE \cite{devise}& 0.8 & 2.7 & 7.9 & 14.2&22.7 \\
 & f-VAEWGAN-D2* \cite{DBLP:conf/cvpr/XianSSA19}&4.76 & 12.68& 19.11& 29.59& 42.89 \\
     & Gvf  \cite{DBLP:conf/iccvw/JurieBH17}
     & \textbf{4.93} & \textbf{13.02} & 20.81 & 31.48 & 45.31 \\
     &  \textbf{CM-GAN} & 0.18 & 7.06 & \textbf{22.55} & \textbf{34.17} & \textbf{46.86}  \\
     
    \bottomrule

    \end{tabular}    
    \endgroup
    
    \caption{Generalized Zero-Shot Learning results on ImageNet-Full. Models marked with * were re-implemented.}
    \label{zsl_imagenet}
\end{table}

% PRESENTATIONS RESULTATS
Quantitative results on ImageNet-Full are provided in \autoref{zsl_imagenet} for the ZSL and G-ZSL tasks, and results on ImageNet-360 are provided in \autoref{360} for ZSL. % --- as done in the literature, only the $R_5$ metric is reported.

% COMMENTAIRES SUR LE ZSL (FULL + 360)
For ZSL, CM-GAN generally outperforms all models on \textit{All}, \textit{3-hop} and \textit{2-hop}  --- results on these benchmarks are increasing (\textit{All}$<$\textit{3-hop}$<$\textit{2-hop}) due to the rising visual and semantic proximity of the class labels to the seen class labels, confirming previous works findings. 
We notice that Gvf, which generates novel visual exemplars, shows better performances than models that learn linear or non-linear cross-modal projections (DeViSE, EXEM) or hybrid models such as CONSE or SYNC. 
On ImageNet-360, CM-GAN outperforms methods than rely on generative models and VAEs such as VZSL, CVAE-ZSL and SE-ZSL, thus proving that our approach based on CycleGAN captures interesting information from the distribution of unseen classes. 
We observe that CM-GAN outperforms DIPL* on ImageNet-Full, and DIPL outperforms CM-GAN on ImageNet-360. Indeed, DIPL's transductive loss aims at bringing closer visual features to the closest class label among unseen classes: while this method efficiently constrains the solution when the number of unseen classes is low (360), its nearest neighbor search over a large number of classes (20K) may bring additional noise that degrades ZSL results. 

% COMMENTAIRES SUR GZSL
Comparing G-ZSL to ZSL allows to analyze whether a model has a tendency to predict seen classes first. For G-ZSL, because they are based on nearest neighbor search, CONSE, CONSE* and CM-GAN perform worse than Gvf (a classifier over seen and unseen classes) at low recall ranks. Interestingly, our re-implementation CONSE*, which is based on a better CNN than the original CONSE, is even more penalized. At higher ranks, we notice that CM-GAN eventually outperforms Gvf for G-ZSL, showing that the cross-modal functions are correctly learned, albeit with a slight overfit towards seen classes. This confirms findings made in \cite{DBLP:journals/pami/XianLSA19} for CONSE, which shares the same results tendencies than CM-GAN --- this result is logic as our model builds upon CONSE (\autoref{conse}). 

\begin{table}[t]                                                                   
    \centering 
        \begin{tabular}{l c}
        \toprule
            Model & $\text{FH}_{5}$\\
    \hline

    DeViSE \cite{devise}& 12.8\\ 
  %  AMP \cite{DBLP:conf/cvpr/FuXKG15} & 13.1 \\
    ConSE \cite{DBLP:journals/corr/NorouziMBSSFCD13} & 15.5\\
 %   SS-Voc \cite{DBLP:conf/cvpr/FuS16}&  16.8\\
    VZSL \cite{DBLP:conf/aaai/WangPVFZCRC18}& 23.1 \\
    CVAE-ZSL \cite{DBLP:conf/cvpr/MishraRMM18}&  24.7 \\
    SE-ZSL  \cite{DBLP:conf/cvpr/VermaAMR18}&  25.4 \\
  %  DEM \cite{DBLP:conf/cvpr/ZhangXG17} & 25.7 \\
  % SAE  \cite{DBLP:conf/cvpr/KodirovXG17} & 27.2 \\
  % JCMSPL \cite{DBLP:journals/corr/abs-1906-05879}& 27.5 \\
  % LESAE \cite{DBLP:conf/ijcai/LiuGLH018}& 27.6 \\
   %  SFS \cite{DBLP:journals/corr/abs-1810-08332} & 28.2\\
     DIPL \cite{DBLP:conf/nips/ZhaoDGLXW18} & \textbf{31.7}\\

    \textbf{CM-GAN} & 25.9\\
    \bottomrule 
    \end{tabular} 

   % \vspace{-0.2cm}
    \caption{ZSL results on ImageNet-360.}
  %  \vspace{-0.2cm}
    \label{360}
\end{table}

\begin{table}[t]                                               \centering 
        \begin{tabular}{l ccc ccc }
        \toprule
            & \multicolumn{3}{c}{ZSL} & \multicolumn{3}{c}{G-ZSL}\\
            \cmidrule(r){2-4}
            \cmidrule(l){5-7}
            Model &  $2_{hop}$ & $3_{hop}$ & All &  $2_{hop}$ & $3_{hop}$ & All\\
        \hline 
    Random &  50 & 50 & 50 & 50 & 50 & 50\\
Init. \phantom{aa} CONSE &   8.88 &12.37 & 15.35 & 9.88 & 12.71 & 15.65 \\
   cycle \phantom{aaa}$\mathcal{L}_c$ &7.32 & 10.84   &13.87& 7.81 & 11. &14.3\\ %comp/ucmr__dataset_ImageNet__multi_m0__lambda_cycle_gan_1__max_steps_2400/8qm2v46hs6
gan \phantom{aaa} $\mathcal{L}_{gan}$    & 7.51& 11.41& 14.25& 7.57 & 11.03&  13.68 \\
  cgan \phantom{aa} $\mathcal{L}_{gan}+\lambda_c.\mathcal{L}_c$ %\hspace{-0.1cm}+\hspace{-0.1cm}\lambda_c.\mathcal{L}_c$
   & 7.34&10.73 &  13.34& 7.72 &10.79 &  13.32 \\

sup \hphantom{aaa} $\mathcal{L}_{sup}$  &   5.83& 8.88  & 11.07 &  6.06 & 8.87 & 11.13 \\
  %  No Replac. & 5.83& 8.88  & 11.07 &  6.06 & 8.87 & 11.13 \\
   % sup + sup & \\  %comp/ucmr__dataset_ImageNet__multi_m1__lambda_supervised_1__max_steps_10000/iab479gnxp
%$ \text{CM-GAN}_{\lambda_c=0}$    & 5.82 & 9.69& & 5.83& 9.57\\% & 5.99& 9.68 & 12.37 &6.21& 9.67 &12.41\\
    %comp/ucmr__dataset_ImageNet__multi_m3__identity_True__max_steps_0__nb_layer_1/n9hy695bgx
    \hline 
      \textbf{CM-GAN} &   \textbf{5.17} &  \textbf{8.23} &  \textbf{9.98}&  \textbf{5.3} & \textbf{8.17} & \textbf{9.97} \\
  %  \hline
 %   Init. & 9.88 & 12.71 & 15.65 \\
  %  cycle & 7.81 & &14.3\\ %comp/ucmr__dataset_ImageNet__multi_m0__lambda_cycle_gan_1__max_steps_2400/g9nz5505a0 
  %  c-gan & 7.72 &10.79 &  13.32 \\
  %  gan  & 7.57 & 11.03&  13.68 \\
  %  sup&  6.06 & 8.87 & 11.13 \\
  %  sup + sup & \\ %comp/ucmr__dataset_ImageNet__multi_m1__lambda_supervised_1__max_steps_10000/6tqml0r24t
  %  Fin ($\lambda_c=0$) & \\ %comp/ucmr__dataset_ImageNet__multi_m3__identity_True__max_steps_0__nb_layer_1/qu1s62h84r
  %  Fin  &  \textbf{5.3} & \textbf{8.17} & \textbf{9.97} \\
  \bottomrule
    \end{tabular} 
    \caption{Ablation study. Measure: MFR (the lower the better).  }
    \label{sup_vs_unsup}
\end{table}

%Results on ImageNet-360 are provided in Table  \ref{360}. As done in the literature, we only report the $F_5$ metric. 
%a subset of 360 target classes: this setting is widely adopted nowadays \cite{DBLP:conf/cvpr/FuS16,DBLP:conf/aaai/WangPVFZCRC18,DBLP:conf/cvpr/MishraRMM18,DBLP:conf/cvpr/VermaAMR18} because evaluation is much faster. 
%We observe that:
%(i) 
%CM-GAN, based on CycleGAN, outperforms methods than rely on generative models and VAEs such as VZSL, CVAE-ZSL and SE-ZSL, 
%(ii) DIPL performs better than CM-GAN, probably because their transductive loss is better able to constrain the solution. However, when the number of classes is higher e.g, on ImageNet-Full, we show in Table \ref{zsl_imagenet} that CM-GAN outperforms our re-implementation of DIPL. Indeed, DIPL's transductive loss aims at bringing closer visual features to the closest class label among unseen classes, showing that a search over low number of classes (360) is more insightful than a search over large number of classes (20K), which only brings noise.

\begin{figure}\centering\includegraphics[width = 0.6\columnwidth]{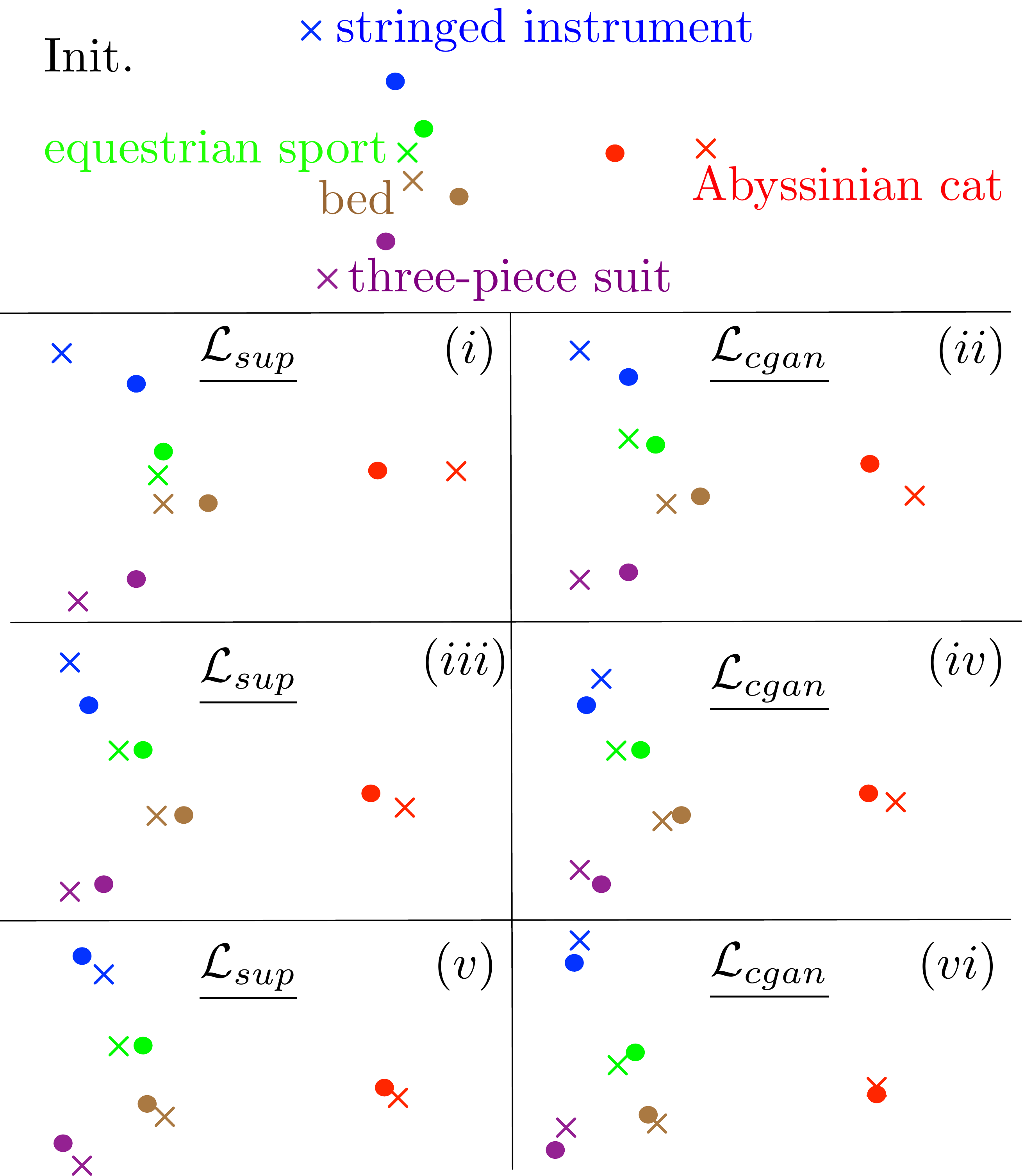}\caption{PCA visualization of visual and textual representations (represented in the same space) along with model's iterations for five randomly sampled unseen classes. Circle: centroid of visual features. Cross: class label. Between even and uneven (resp.\ uneven and even) steps, visual centroids (resp.\ class labels) are moving.  %Explained variance ratio: [0.19, 0.13].
}\label{pca1}\end{figure}

To further analyze our model, we provide an ablation study in \autoref{sup_vs_unsup}. We report models where losses $\mathcal{L}_{c}$, $\mathcal{L}_{gan}+\lambda_c.\mathcal{L}_c$, $\mathcal{L}_{gan}$ and $\mathcal{L}_{sup}$ are optimized individually. Init. corresponds to the initialization of our model, with CONSE embeddings as visual features (i.e. $V_0$ and $T_0$). 
We observe that:
(1) $\text{sup}>\text{cgan}$: the supervised step is better than the transductive step, which is expected since the alignment information is used;
(2) $\text{cgan}>\text{gan}$ and $\text{cgan}>\text{cycle}$, which indicates that both cycle-consistency and adversarial losses are complementary; 
(3) $\text{CM-GAN} > \text{sup}$ and $\text{CM-GAN} >\text{cgan}$, which shows the benefits of the alternating learning process; %, except when $\lambda_c$ since we have $\text{CM-GAN}(\lambda_c=0)<\text{sup}$;
(4) Our final model CM-GAN is the best scenario. 
The complementarity of supervised and transductive objectives is illustrated in \autoref{pca1}: we randomly sample five unseen classes and visualize the evolution of their textual and visual representations at each model optimization step (\autoref{learning}). We observe that modalities align with each supervised (i,iii,v) and unsupervised (ii,iv,vi) iteration. 
%Furthermore, Figure \ref{mfr_time}  illustrates the evolution of the MFR measure in the course of training for the ZSL and G-ZSL tasks. We observe that (1) the first sup. step gives the strongest improvements, (2) the improvement is decreasing step after step until converging at step 6.  \begin{figure*} \centering \includegraphics[width = 0.8\textwidth,trim={0 0 8cm 0},clip]{mfr.pdf} \caption{Mean First Rank over model iterations. Left (resp.right): ZSL (resp. G-ZSL) task. } \label{mfr_time} \end{figure*}

%We also provide a fine-grained evaluation in Table \ref{finegrained}. As done in \cite{DBLP:journals/pami/XianLSA19}, we distinguish various set of unseen classes: the set of $n$ most (resp. least) populated classes $M_n$ (resp. $L_n$) with $n \in \{500,1000,5000\}$.      \input{finegrained.tex}

%\vspace{-0.2cm}

%\vspace{-0.5cm}

%\paragraph{How does CONSE+sup compare with CONSE + sup + sup ? (Figure \ref{supsup})}
%\begin{figure} \centering \includegraphics[width = \columnwidth]{conse_sup_sup.pdf} \caption{CONSE+sup vs CONSE+sup+sup.} \label{supsup} \end{figure}
%No difference (the same information is used redundantly).
%\paragraph{What is the impact of a Cycle-Gan loss on another model than CONSE, namely DEVISE ? (Figure \ref{devise})}
%\begin{figure*} \centering \includegraphics[width= \textwidth]{devise.pdf} \caption{CycleGan applied to the DeVISE model.} \label{devise}\end{figure*}
%The generator and discriminator losses converge to 0. Retrieval scores are not improved.

%In Table \ref{dipl}, we see that DIPL outperforms our model on ImageNet-360. However, on ImageNet-Full, our model outperforms our re-implementation of DIPL (except for the MFR metric). This could be explained by the fact that DIPL leads to a low MFR; but, for a very wide class corpus such as the ImageNet-Full, the MFR is not so important to get high $F_i$ scores for low values of $i$; on the contrary, in the 360 setting, a high MFR is more likely to lead to high scores for $F_i$.  \input{dipl.tex}

%\begin{figure} \centering \includegraphics[width = \columnwidth]{quali.pdf}\caption{PCA Visualization of the various stages of our model on 5 randomly selected classes from $\text{Most}_{500}$. Cross: centroid of image rep. Circle: word rep.} \label{pca2} \end{figure}

\subsection{Learning Grounded Word Representations with CM-GAN}
\label{grounding}

%Language grounding \cite{Lazaridou2015b,collell2017,zablockiaaai2018} is a field aimed at enhancing textual representations using visual information. %

%In this section, we provide a post-hoc analysis of \autoref{expes} and we further validate our approach by using CM-GAN to \textit{visually ground language} \cite{Lazaridou2015b}. More precisely, we study the quality of word embeddings learned in \autoref{expes}, and we investigate whether they capture visual information and common-sense knowledge. 
In this section, we provide a post-analysis of \autoref{expes} and further validate CM-GAN by using it to \textit{visually ground language} \cite{Lazaridou2015b}. More precisely, we study the word embeddings learned in \autoref{expes}, to investigate whether they capture visual information and common-sense knowledge. 

The goal of \textit{grounding} is to enhance textual representations using visual information. 
The first motivation for language grounding came from cognitive psychology studies \cite{80449942ebb84873aef804a1799cd546,arahoafa,ouigoig,gouigoig}, stating that the human understanding of language is intrinsically linked to our perceptual and sensory-motor experience. It was further consolidated by quantitative-based findings such as the Human Reporting Bias \cite{Gordon2013a,DBLP:conf/cvpr/MisraZMG16}: \textit{the frequency at which objects, relations, or events occur in natural language are significantly different from their real-world frequency} --- in other terms, what is obvious (visual commonsense) is rarely stated explicitly in text, since it is supposed to be known by the human reader. Following these considerations, AI researchers began to point out the lack of grounding \cite{DBLP:journals/llc/Baroni16} of Distributional Semantic Models such as Word2Vec \cite{DBLP:conf/nips/MikolovSCCD13}, where the meaning of words only depends on textual co-occurrence patterns. 

Unlike previous grounding models, that leverage direct correspondence between words and images , CM-GAN also exploits unsupervised information: it is thus interesting to determine whether this can enhance our word representations.  
Based on our CM-GAN model trained on ImageNet (\autoref{expes}), we learn grounded word representations using information from seen and unseen classes. 
\cite{collell2017} has shown that a word representation $X$ given by Word2Vec (\textit{purely textual} as it is learned using Word2Vec on textual corpus only) and its projection in a visual space (which is said to be \textit{grounded} since visual information has been incorporated) have complementary information, that can be exploited by concatenating them. Thus, we also evaluate various concatenation combinations between $X$ and its grounded projections.

Evaluations are reported in \autoref{word_300} for standard word embeddings benchmarks \cite{zablockiaaai2018}. First, we consider \textit{semantic relatedness} benchmarks, such as WordSim353 \cite{DBLP:journals/tois/FinkelsteinGMRSWR02}, MEN \cite{DBLP:journals/jair/BruniTB14}, SimLex-999 \cite{DBLP:journals/coling/HillRK15}, SemSim and VisSim \cite{DBLP:conf/acl/SilbererL14}, which give human judgements of similarity (e.g., between $0$ and $10$) for a series of word pairs. For each benchmark, the reported result is the Spearman correlation computed between (i) human similarity scores and (ii) the cosine similarities between word vectors. 
Second, we report results for the \textit{concreteness prediction} task on the USF dataset \cite{Nelson2004}, which gives concreteness scores for 3260 English words. A SVM model is trained to predict these scores from the word vector. The reported result is the accuracy on a test set.

%Results are reported in Table \ref{word_300} the grounded vectors $V_{\text{sup}}(X)$ and $V_{\text{uns}}(X)$, along with their concatenation with $X$ itself, which according to \cite{collell2017} should integrate complementary information from both modalities. 
%Following \cite{collell2017}, we also evaluate variants where we concatenate the original Word2vec vectors with our grounded vectors, that we re-project in dimension $300$ using a PCA to allow comparison with equal dimensions.

We use the original Word2Vec vectors (dimension $d=300$) as $T_0$ representations.
We note $V_{\text{sup}}(X)$ (resp.\ $V_{\text{trans}}(X)$) the cross-modal projection of $X$ learned using a supervised (resp.\ transductive) step. 
We observe that grounded vectors tend to outperform $X$ on all benchmarks except Concreteness, and that the information present in $X$ and the grounded vectors is complementary, as $X\oplus V_{\text{trans}}(X)$ (resp.\ $X\oplus V_{\text{sup}}(X)$) outperforms both $X$ and $V_{\text{trans}}(X)$ (resp.\ $V_{\text{sup}}(X)$). Interestingly, we notice that $X \oplus V_{\text{trans}}(X)$ gives the highest performance, thus showing the efficiency of exploiting unsupervised visual information using CM-GAN. 
\newcolumntype{R}[2]{%
    >{\adjustbox{angle=#1,lap=\width-(#2)}\bgroup}%
    l%
    <{\egroup}%
}
\newcommand*\rot{\multicolumn{1}{R{30}{0.8em}}}

\begin{table}[htb]    
%\vspace{0.4cm}
    \centering 
        \begin{tabularx}{\textwidth}{@{}l X X X X X X X@{}}
    \toprule
            Model & \rot{MEN}  & \rot{SemSim} & \rot{SimLex} & \rot{VisSim} & \rot{WordSim} &  \rot{Conc.} & \rot{Avg.} \\ 
    \hline 
    $X$ & 68 & 60 & 33 & 49 & \textbf{62} &  63& 55.8\\ 
    $V_{\text{sup}}(X)$ & 69 & 66 & 34 & 57 & 59 &  59 & 57.3\\
    $X \oplus V_{\text{sup}}(X)$ & 70 & \textbf{69} & 36 & \textbf{58} & 59 & 58 & 58.3 \\
    $V_{\text{trans}}(X)$ & 70 & 62 & 35 & 51 &\textbf{62}&  62& 57 \\
    $X \oplus V_{\text{trans}}(X)$ & \textbf{72} & \textbf{69} & \textbf{37} & 55 & 58 &  \textbf{64}& \textbf{59.2} \\
   % $\text{sup}(X)+\text{cgan}(X)$& 70 & 66 & 35 & 57 & 60 &  53 & 56.8\\
  %  $X \oplus (\text{sup}(X)+\text{cgan}(X))$& 71 & \textbf{69}& 36 & \textbf{58} & 60 &  52& 57.7 \\

    $V_{\text{sup}}(X)\oplus V_{\text{trans}}(X)\hspace{-0.2cm}$ & 70 & \textbf{69} & 36 & \textbf{58} & 59 & 58 & 58.3 \\

  %  $X \oplus \text{sup}(X) \oplus \text{cgan}(X)$& 70 &\textbf{69} & 36 & \textbf{58} & 60 & 57& 58.3 \\
    \bottomrule 
    \end{tabularx} 
    %\vspace{-0.1cm}
    \caption{Grounded vectors evaluation. Concatenated vectors were projected with PCA so that all vectors have the same dimension. }
    \label{word_300}
\end{table}

%\paragraph{Experiment 2}
%In a second experiment (\autoref{word_500}), we consider the functions learned in \autoref{expes}. TODO
%\input{word_500_2.tex} 

\subsection{Zero-Shot Sentence-to-Image Matching}
\label{section_mscoco}

\begin{table}[t]                                   
    \centering 
        \begin{tabular}{l cccc cccc  }
        
        \toprule

        & \multicolumn{4}{c}{Text to Image}&  \multicolumn{4}{c}{Image to Text}\\
       \cmidrule(r){2-5}
        \cmidrule(l){6-9}
        
        Model  & MFR & $FH_{1}$ &$FH_{5}$ & $FH_{10}$ & MFR & $FH_{1}$ &$FH_{5}$ & $FH_{10}$ \\  
    \hline 
    Rand. & 50 & 0.1 & 0.5& 1 & 50 & 0.1 & 0.5& 1\\ 
    CONSE & 20.3  & 3.3 & 13.9 & 23.5& 14.1 & 2.8& 12.3 & 20.2\\ 
    $\mathcal{L}_{gan}$ & 21 & 4.2 & 15.7 & 25.2 & 14.1 & 3.1 & 15.1 & 23.7  \\
    $\mathcal{L}_{c}$& 18.3 & 4.7 & 14.3& 24 & 11.3 & 3.7 & \textbf{16.6} & 27.3\\
    $\mathcal{L}_{cgan}$ & \textbf{15.7} & \textbf{5.7} & \textbf{17.2} & \textbf{27.7}& \textbf{11.1}& \textbf{4.1} & 15.5 & \textbf{27.7}\\[0.25cm]
   % cgan ($10)$ & 16.2 & 5 & 16.8& 26.6 &12.2 &3.6 & 15.5 & 26\\ 
%    cgan ($10^2)$& 17.8 & 5.2 & 15.7 & 26  & 13.1 &2.9 & 12 & 21.7 \\ 
 %   cgan ($10^3)$ &17.3 & 5.1 & 15.4 & 24.8& 12 & 3.4 & 15.7 & 26.2\\ 

    $\mathcal{L}_{sup}$  &  \textit{7.3} & \textit{8.3} & \textit{26.3} & \textit{39.8}  & \textit{4.5} & \textit{7.1} & \textit{28.1} & \textit{44.7}\\
     \bottomrule 

    \end{tabular} 
    \caption{Cross-modal retrieval results on MS COCO.}
    \label{mscoco}
\end{table}

In this section, we demonstrate that CycleGan can align the visual and textual modalities \textit{without any supervision}. We also show that CM-GAN can be applied to other settings than ImageNet: (i) replacing words by sentences, and (ii) with as many classes as training examples (590K sentences). % --- generatlization is made possible here a reasonable number of classes (20K) by a high number of sentences (590K) --- which share common words so that generalization is possible. 

We evaluate several scenarios of our model on the standard cross-modal retrieval task: given a sentence, retrieving the closest image (Text to Image) and vice versa (Image to Text), in \autoref{mscoco}. 
We observe that (1) the initialization (CONSE) already shows substantial improvement compared to the Random baseline, (2) models based on CycleGAN improve performances compared to the CONSE model, due to the beneficial action of the CycleGAN loss, (3) CM-GAN has much lower performance than the Supervised baseline (where $\mathcal{L}_{sup}$ is learned), which is intuitively expected. %since many words can re-appear in many sentences. 

%\paragraph{Experiment 1 (Table \ref{word_500})} At each step of our model, a function $V$ is learned from the textual space $\mathcal{T}$ to the visual space $\mathcal{V}$. At step 1, since the supervised loss is optimized, we call $G=\text{sup}_1$; at step 2, since the Cycle-Gan loss is optimized, we call $G=\text{cgan}_2$, etc. \footnote{Since we use the word embeddings of \cite{DBLP:conf/cvpr/ChangpinyoCGS16} (for a fair comparison against the literature) for the 21K classes of ImageNet, and the authors do not grant access to the whole Word2vec model, we learned a Word2vec model, and learned a projection between our model and theirs on the 21K ImageNet classes. We call $X$ this projection, which is thus very close to their original Word2vec model.}  \input{word_500.tex}

\section{Conclusion and Future Works}

In this paper, we propose the CM-GAN model for T-ZSL. 
We demonstrate that:
(1) CM-GAN is successful on the ImageNet T-ZSL task, with state-of-the-art results, 
(2) visual and textual modalities can be somewhat aligned without supervision on a zero-shot sentence-to-image task on MS COCO, 
(3) textual representations can be enhanced using CM-GAN. 

We conclude that there is meaningful information to be exploited in the similarities between textual and visual distributions, even in cases where there is no direct text/vision supervision. We showed it for three multimodal tasks: Transductive Zero-Shot Learning, Zero-Shot Image-to-Sentence Matching and Visual Grounding of Language.

%As future work, we would study the slight over-fitting of CM-GAN for G-ZSL, and investigate the application of CM-GAN to other configurations, such as speech/text or text/video. 

We now present research perspectives.

%\paragraph{Zero-Shot Sentence-to-Image Matching}In this paper, we performed zero-shot sentence-to-image matching by training a Cross-Modal CycleGAN model on MS COCO data. We represented a sentence by the sum of its word embeddings, and we represented images using a convex combination of the most probable class labels embeddings, with the CONSE model \citep{DBLP:journals/corr/NorouziMBSSFCD13}. We showed that the CycleGAN model could, without any supervision, capture some text/vision alignment; however, improvements compared to the CONSE initialization could be done with a new method to encode sentence, able to take into account word order. 
First, to improve performances on the Zero-Shot Sentence-to-Image Matching task, an interesting perspective would be to encode sentences and images using recent improvements on Multimodal Language Models, such as LXMERT \cite{DBLP:conf/emnlp/TanB19} or VisualBERT \cite{DBLP:journals/corr/abs-1908-03557}. Indeed, these models enable to produce cross-modal representations that could be integrated within our Cross-Modal model. 

Another interesting perspective would be to consider ZSL learning with noisy text descriptions, instead of class labels.
Even though ZSL is mostly focused on classifying images with class labels (e.g, \textit{Abyssinian cat}, \textit{dog}, \textit{traffic light}), some works \citep{DBLP:conf/cvpr/ElhoseinyZZE17,DBLP:conf/cvpr/ZhuEL0E18} have proposed, instead, to consider \textit{noisy} text descriptions e.g, \textit{The Parakeet Auklet is a small (23cm) auk with a short orange bill that is upturned ...}  We could apply our CrossModal CycleGAN model to this task: to do so, we would need to  select useful information in the text (visual words like \textit{small} or \textit{orange} for example) and represent the text as a weighted linear combination of these salient words. We could even consider a fully unsupervised configuration (as done in the zero-shot text-to-image matching task) in which we do not use a direct supervision between images and textual descriptions.

\bibliography{main}

\end{document}